\documentclass[conference,a4paper]{IEEEtran}
\IEEEoverridecommandlockouts
\usepackage{cite}
\usepackage{amsmath,amssymb,amsfonts}
\usepackage{algorithmic}
\usepackage{graphicx}
\usepackage{textcomp}
\usepackage{xcolor}
\usepackage{tikz}
\usepackage{float}
\usepackage{pgf-pie}
\usepackage{pgfplots}
\usepackage{balance}
\usetikzlibrary{positioning, decorations.pathreplacing}
\pgfplotsset{compat=1.18}
\def\BibTeX{{\rm B\kern-.05em{\sc i\kern-.025em b}\kern-.08em
    T\kern-.1667em\lower.7ex\hbox{E}\kern-.125emX}}

\begin{document}

\title{Large-Scale Evaluation of Advanced Imputation Methods for Missing Values in Smart Meter Data}

\author{
\IEEEauthorblockN{Daniela Stojcheska}
\IEEEauthorblockA{\textit{FEEIT} \\
\textit{Ss. Cyril and Methodius University in Skopje}\\
Skopje, North Macedonia \\
danielas@feit.ukim.edu.mk}
\and
\IEEEauthorblockN{Marija Markovska}
\IEEEauthorblockA{\textit{FEEIT} \\
\textit{Ss. Cyril and Methodius University in Skopje}\\
Skopje, North Macedonia \\
marijam@feit.ukim.edu.mk}
\and
\IEEEauthorblockN{Dimitar Taskovski}
\IEEEauthorblockA{\textit{FEEIT} \\
\textit{Ss. Cyril and Methodius University in Skopje}\\
Skopje, North Macedonia \\
dtaskov@feit.ukim.edu.mk}
\and
\IEEEauthorblockN{Branislav Gerazov}
\IEEEauthorblockA{\textit{FEEIT} \\
\textit{Ss. Cyril and Methodius University in Skopje}\\
Skopje, North Macedonia \\
gerazov@feit.ukim.edu.mk}
\and
\IEEEauthorblockN{Boris Nikolov}
\IEEEauthorblockA{\textit{Elektrodistribucija DOOEL} \\
\textit{EVN Group}\\
Skopje, North Macedonia \\
boris.nikolov@dso.mk}
}

\IEEEpubid{\makebox[\columnwidth]{\fontsize{9}{11}\selectfont
979-8-3195-1058-7/26/\$31.00~\copyright2026 IEEE\hfill}
\hspace{\columnsep}\makebox[\columnwidth]{}}
\maketitle

\begin{abstract}
Accurate and reliable collection of electricity consumption data through Advanced Metering Infrastructure (AMI) is of great importance for the operation of smart grids, especially for the detection of non-technical losses (NTL). However, real-world datasets frequently suffer from missing values due to communication failures. This paper presents an empirical evaluation of three advanced algorithms for large-scale data imputation: the Optimally Weighted Average (OWA) method, Low-Rank Matrix Completion via SoftImpute, and a Shape-Modeling Autoencoder. Existing studies on missing value imputation in electricity consumption data often lack validation on larger datasets. Therefore, the goal of this paper is to validate the selected algorithms on a large-scale real-world electricity consumption dataset from North Macedonia that includes 17,428 commercial smart meters over two years. The robustness of each algorithm is evaluated by simulating continuous gaps in the data ranging from 1 to 168 hours. The results indicate that OWA provides the lowest overall reconstruction error across the evaluated gap sizes and strong stability in worst-case scenarios for gaps of up to one week. In contrast, the autoencoder exhibits higher variance, while SoftImpute has stable but inferior accuracy. These findings suggest that imputation methods should be selected based on the characteristics of load curve data and highlight the potential for hybrid algorithmic architectures in future grid management systems.
\end{abstract}

\renewcommand\IEEEkeywordsname{Keywords}
\begin{IEEEkeywords}
smart meter data, data imputation, Optimally Weighted Average (OWA), low-rank matrix completion, autoencoders, non-technical loss (NTL)
\end{IEEEkeywords}

\section{Introduction}
The widespread deployment of Advanced Metering Infrastructure (AMI) provides utility companies with high-resolution time series data, enabling advanced operations such as load forecasting and structural grid monitoring. Crucially, these data are the foundational pillar for feature extraction and detection of Non-Technical Losses (NTL), which include electricity theft and billing fraud \cite{markovska_tcn, bilbiloska_feat}.

However, real-world AMI networks are highly susceptible to data loss caused by hardware malfunctions and communication bottlenecks. These outages range from short one-hour breaks to long multi-day outages. Traditional data handling approaches, such as simple interpolation, severely distort the natural consumption profile when applied to larger gaps. In NTL detection, such distortions substantially degrade algorithmic performance, leading to false positive theft alarms or masking genuine illegal consumers.

Therefore, advanced data imputation techniques are required to accurately reconstruct missing segments. Although there are various complex algorithms, they are predominantly evaluated on localized datasets (e.g., a single building or a few hundred meters) or synthetically generated load profiles \cite{peppanen_imputation, mateos_lowrank, duarte_autoencoder, kim_chillers}. Their practical scalability and worst-case robustness on massive utility networks remain largely unexplored.

This paper addresses the aforementioned significant gap by benchmarking three distinct data imputation paradigms on a highly diverse, real-world dataset of 17,428 active commercial smart meters from North Macedonia. By rigorously analyzing performance across gap sizes from 1 to 168 hours, the practical limitations of each method are identified, and a practical empirical baseline for real-world AMI imputation is established.

\section{Related Work}
Missing data imputation in time series analysis has evolved from classical statistical methods to machine learning architectures. Recent reviews strongly highlight the necessity of robust imputation specifically tailored for smart grids  \cite{schreiber_review}. 

Various approaches have been proposed, including bagging ensembles of Multilayer Perceptrons \cite{jung_mlp}, Denoising Autoencoders \cite{ryu_dae}, domain-specific K-Nearest Neighbors \cite{kim_chillers}, and the integration of Time Series Foundation Models \cite{sartipi_foundation} or advanced Transformers like TimesNet \cite{gao_timesnet}. Rather than proposing a single new model, this study comprehensively benchmarks three established algorithms representing fundamentally different imputation approaches on a large-scale, real-world AMI dataset:
\begin{enumerate}
    \item \textbf{Temporal Dependency:} The Optimally Weighted Average (OWA) \cite{peppanen_imputation}, utilizing local historical and future context.
    \item \textbf{Spatial Correlation:} Low-Rank Matrix Completion (SoftImpute) \cite{mateos_lowrank}, leveraging global network similarities.
    \item \textbf{Deep Learning:} A Shape-Modeling Autoencoder \cite{duarte_autoencoder}, learning nonlinear daily profile shapes.
\end{enumerate}

The vast majority of existing literature evaluates these algorithms on highly restricted datasets \cite{peppanen_imputation, kim_chillers} or relies on artificial anomalies \cite{mateos_lowrank, sartipi_foundation}. By testing these models across a continuous 25-month period on over 17,000 real commercial meters, this study provides a much-needed, large-scale empirical validation of their performance during extreme data outages.

\section{Dataset and Preprocessing}

\subsection{Raw Data and Aggregation}
The empirical evaluation is based on a real large-scale smart meter dataset provided by Elektrodistribucija DOOEL, the electricity distribution system operator in North Macedonia. The dataset exclusively contains nonresidential commercial and industrial meters, which exhibit significantly more erratic consumption behaviors than standard households. The raw data span 25 continuous months (1 June 2023 to 30 June 2025). Initially recorded at a 15-minute resolution, the readings were aggregated into hourly intervals to reduce computational complexity. 

\subsection{Data Cleansing and Filtering}
In practice, smart meter data inherently contains noise, communication failures, and estimates made on the utility’s side. To construct a reliable ground truth, a systematic preprocessing pipeline was designed. All recorded measurements were retained, including verified physical readings, utility-generated estimations, and explicit hardware transmission failures, to preserve structural continuity. Furthermore, missing timestamp rows that were entirely absent from the raw database were injected, labeling them with a custom \texttt{Missing} status. Consequently, every meter's profile was standardized into an unbroken sequence of exactly 18,264 hourly intervals.

To maintain experimental integrity, we removed recently installed meters, inactive connections ($<100$ kWh cumulatively), and meters with continuous outages exceeding one week. This filtering yielded a reliable dataset of 17,428 active smart meters.

\subsection{Missingness Profile and Load Volatility}
To fully understand the imputation challenge, the structural characteristics of the working dataset must be analysed. An evaluation of the cleansed data reveals the severity of the missing data problem. Not a single meter ($0\%$) in the dataset possessed a perfect ground truth record; all 17,428 meters experienced data gaps, with individual losses ranging from a minimum of just 2 hours to a maximum of 168 hours (one full week). This widespread deficiency resulted in a cumulative network loss of 651,652 hours over the 25 months. Table \ref{tab:missingness} illustrates the distribution of missing data across the network.

\begin{table}[!b]
\caption{Distribution of Missing Data Across the Meter Network}
\label{tab:missingness}
\begin{center}
\begin{tabular}{|c|c|c|}
\hline
\textbf{Missing Gap Percentage} & \textbf{Number of Meters} & \textbf{Percentage} \\
\hline
$< 0.1\%$ ($< 18$ hours) & 7,545 & 43.29\% \\
\hline
$0.1\% - 0.5\%$ ($18 - 91$ hours) & 8,271 & 47.46\% \\
\hline
$0.5\% - 1.0\%$ ($91 - 182$ hours) & 1,283 & 7.36\% \\
\hline
$1.0\% - 5.0\%$ ($182 - 913$ hours) & 329 & 1.89\% \\
\hline
\end{tabular}
\end{center}
\end{table}

Furthermore, the distribution of the longest continuous gaps per meter is also analyzed. As shown in Table \ref{tab:max_gaps}, the majority of meters exhibit only short outages, with 10,223 meters having maximum continuous gaps shorter than 6 hours. A significant portion of the network (5,922 meters) experiences moderate outages lasting up to 24 hours, while a small group (1,283 meters) is affected by longer continuous gaps lasting up to a week (168 hours). It should be noted that, as a direct result of the applied filtering protocol, no meter exhibits continuous outages exceeding one week, ensuring that the imputation tasks remain within a practically recoverable algorithmic range.

\begin{table}[!b]
\caption{Distribution of Meters by Maximum Continuous Gap Length}
\label{tab:max_gaps}
\begin{center}
\begin{tabular}{|c|c|c|}
\hline
\textbf{Maximum Continuous Gap} & \textbf{Number of Meters} & \textbf{Percentage} \\
\hline
$\le 6$ hours & 10,223 & 58.66\% \\
\hline
$6 - 24$ hours & 5,922 & 33.98\% \\
\hline
$24 - 168$ hours (1 week) & 1,283 & 7.36\% \\
\hline
\end{tabular}
\end{center}
\end{table}

\begin{figure}[!b]
    \centering
    \begin{tikzpicture}
        \pie[
            text=legend, 
            radius=1.7, 
            color={green!30, yellow!40, red!40}
        ]{
            18.90/Stable ($CV \le 0.5$), 
            24.19/Moderate ($0.5 < CV \le 1.0$), 
            56.91/Highly Volatile ($CV > 1.0$)
        }
    \end{tikzpicture}
    \caption{Categorization of the smart meter network based on load volatility (Coefficient of Variation).}
    \label{fig:cv_pie}
\end{figure}
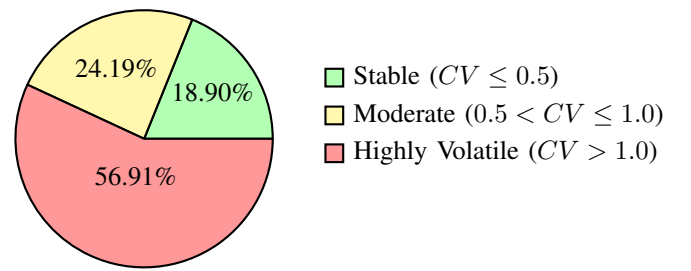

Beyond the frequency of missing values, the fundamental challenge lies in the unpredictability of the consumption patterns. The Coefficient of Variation (CV) measures the relative dispersion of consumption data. As illustrated in Fig.~\ref{fig:cv_pie}, over half of the commercial meters ($56.91\%$) are categorized as highly volatile ($CV > 1.0$). This volatility is typically driven by strict operational schedules, such as commercial facilities operating exclusively from 8 AM to 4 PM, creating extreme daily spikes followed by near-zero nighttime consumption.

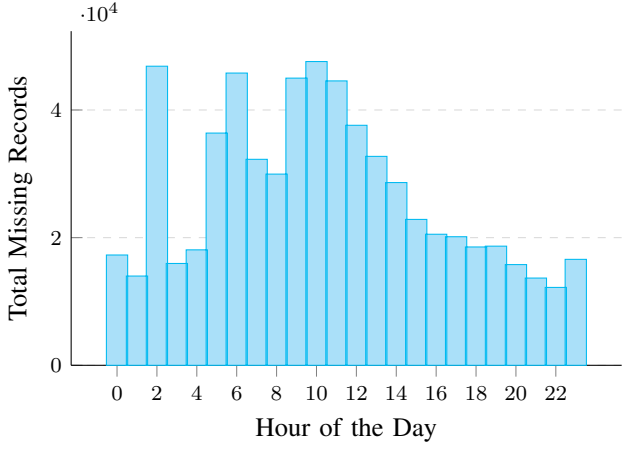
\begin{figure}[t]
    \centering
    \begin{tikzpicture}
        \begin{axis}[
            width=\columnwidth,
            height=6cm,
            ybar,
            bar width=8pt,
            xlabel={Hour of the Day},
            ylabel={Total Missing Records},
            ymin=0,
            xtick={0,2,4,6,8,10,12,14,16,18,20,22},
            tick label style={font=\footnotesize},
            ymajorgrids=true, 
            grid style={dashed, gray!30},
            axis x line*=bottom,
            axis y line*=left
        ]
        \addplot[fill=cyan!30, draw=cyan!80] coordinates {
            (0,17281) (1,13983) (2,46859) (3,15954) (4,18088) (5,36383)
            (6,45786) (7,32272) (8,29955) (9,44992) (10,47591) (11,44556)
            (12,37603) (13,32733) (14,28630) (15,22857) (16,20532) (17,20144)
            (18,18539) (19,18666) (20,15779) (21,13668) (22,12202) (23,16599)
        };
        \end{axis}
    \end{tikzpicture}
    \vspace{-2mm}
    \caption{Temporal distribution of missing data occurrences by hour of the day.}
    \label{fig:dist_hour}
    \vspace{-3mm}
\end{figure}

\begin{figure}[t]
    \centering
    \begin{tikzpicture}
        \begin{axis}[
            width=\columnwidth,
            height=6cm,
            ybar,
            bar width=14pt,
            xlabel={Month of the Year},
            ylabel={Average Missing Records},
            ymin=0,
            xtick={1,2,3,4,5,6,7,8,9,10,11,12},
            xticklabels={Jan, Feb, Mar, Apr, May, Jun, Jul, Aug, Sep, Oct, Nov, Dec},
            x tick label style={rotate=45, anchor=east, font=\footnotesize},
            y tick label style={font=\footnotesize},
            ymajorgrids=true, 
            grid style={dashed, gray!30},
            axis x line*=bottom,
            axis y line*=left
        ]
        \addplot[fill=cyan!30, draw=cyan!80] coordinates {
            (1,11510) (2,11030) (3,33024) (4,15692) (5,42004) (6,26649)
            (7,29884) (8,22029) (9,21014) (10,34025) (11,40495) (12,25142)
        };
        \end{axis}
    \end{tikzpicture}
    \caption{Normalized distribution of missing data across months.}
    \label{fig:dist_month}
    \vspace{-3mm}
\end{figure}
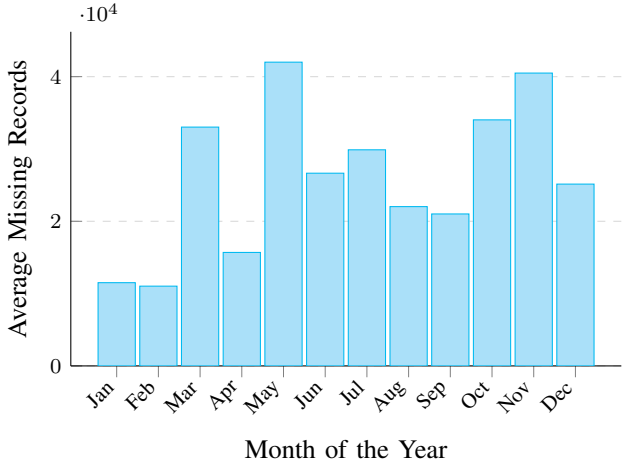

To further understand the nature of these data outages, Fig.~\ref{fig:dist_hour} and Fig.~\ref{fig:dist_month} illustrate how missing values are distributed across hours of the day and months of the year. The results indicate that missing values tend to cluster around specific time periods rather than occurring randomly throughout the day. We observe significant spikes in missing data during the early morning (02:00, 06:00) and right in the middle of peak business hours (09:00 to 11:00). On a seasonal scale, outages occur most frequently in May and November. 

\vspace{-2mm}

\vspace{-2mm}

\section{Methodology of Imputation Algorithms}
This section outlines the underlying logic and mathematical formulations of the three imputation methods under consideration. Because commercial energy consumption is highly unpredictable, this study avoids reliance on standard parameters. Instead, it introduces a data-driven tuning approach designed specifically for the analyzed smart meter network.

\subsection{Optimally Weighted Average (OWA)}
The Optimally Weighted Average (OWA) algorithm, originally proposed by Peppanen et al. \cite{peppanen_imputation}, is a highly effective deterministic method that reconstructs missing data by intelligently blending local sequential context with global historical patterns. 

For any missing hour $i$ within a continuous data gap of length $G$, the OWA estimates the missing consumption $y^{OWA}_i$ as a weighted combination of Linear Interpolation ($y^{LI}_i$) and the Historical Average ($y^{HA}_i$):
\begin{equation}
    y^{OWA}_i = w_i \cdot y^{LI}_i + (1 - w_i) \cdot y^{HA}_i
    \label{eq:owa_base}
\end{equation}

The local component, $y^{LI}_i$, is derived by linearly connecting the last recorded valid measurement immediately before the gap and the first valid measurement immediately after the gap. Conversely, the historical component, $y^{HA}_i$, captures the seasonal and daily operational behavior. This component is calculated by averaging historical values that share the same time-of-day and day-of-week as the missing hour, restricted within a surrounding $\pm 8$-day window to capture local seasonal trends.

The fundamental innovation of OWA lies in its dynamic weighting factor, $w_i$, which dictates the level of trust placed in the linear interpolation. It is governed by an exponential decay function:
\begin{equation}
    w_i = e^{-\alpha \cdot d_i}
    \label{eq:owa_decay}
\end{equation}
where $d_i = \min(i, G - i + 1)$ represents the chronological distance (in hours) from the missing point $i$ to the nearest valid measurement edge, and $\alpha$ is a (positive) weight parameter.

\subsubsection{Data-Driven Alpha Optimization}
To determine the optimal $\alpha$ parameter for each specific gap size, a comprehensive grid search was conducted using a Monte Carlo simulation. We isolated a representative subset of 300 healthy meters and randomly injected 100 continuous outages per meter across 11 different gap sizes ($G \in \{1, 2, 3, 4, 6, 8, 12, 24, 48, 72, 168\}$ hours).

\begin{figure}[tb]
    \centering
    
    \definecolor{sblue}{RGB}{31,119,180}
    \definecolor{sorange}{RGB}{255,127,14}
    \definecolor{sgreen}{RGB}{44,160,44}
    \definecolor{sred}{RGB}{214,39,40}
    \definecolor{spurple}{RGB}{148,103,189}
    \definecolor{sbrown}{RGB}{140,86,75}
    \definecolor{spink}{RGB}{227,119,194}
    \definecolor{sgray}{RGB}{127,127,127}

    \begin{tikzpicture}
        \begin{axis}[
            width=\columnwidth,
            height=6.5cm,
            title={\textbf{Weight function shape for various $\alpha$ parameters}},
            xlabel={$d$ (Distance to nearest known sample in hours)},
            ylabel={$w$ (Weight factor)},
            xmin=0, xmax=20,
            ymin=0, ymax=1.05,
            grid=both,
            grid style={dashed, gray!40},
            legend columns=2,
            legend style={
                at={(0.97, 0.92)}, 
                anchor=north east,
                font=\scriptsize, 
                fill=white, 
                fill opacity=0.9, 
                draw opacity=1
            },
            samples=100,
            very thick,
            no markers
        ]
        
        \addplot[domain=0:20, color=sblue, dashed] {exp(-0.0*x)};
        \addplot[domain=0:20, color=sorange, dashed] {exp(-0.1*x)};
        \addplot[domain=0:20, color=sgreen, dashed] {exp(-0.25*x)};
        \addplot[domain=0:20, color=sred, dashed] {exp(-0.5*x)};
        \addplot[domain=0:20, color=spurple, dashed] {exp(-0.75*x)};
        \addplot[domain=0:20, color=sbrown, dashed] {exp(-1.0*x)};
        \addplot[domain=0:20, color=spink, dashed] {exp(-1.5*x)};
        \addplot[domain=0:20, color=sgray, dashed] {exp(-2.0*x)};
        
        \legend{$\alpha=0.00$, $\alpha=0.10$, $\alpha=0.25$, $\alpha=0.50$, $\alpha=0.75$, $\alpha=1.00$, $\alpha=1.50$, $\alpha=2.00$}
        \end{axis}
    \end{tikzpicture}
    \caption{Exponential decay of the OWA weight factor ($w_i$) as a function of distance $d$ for various $\alpha$ parameters.}
    \label{fig:owa_weights}
    \vspace{-3mm}
\end{figure}
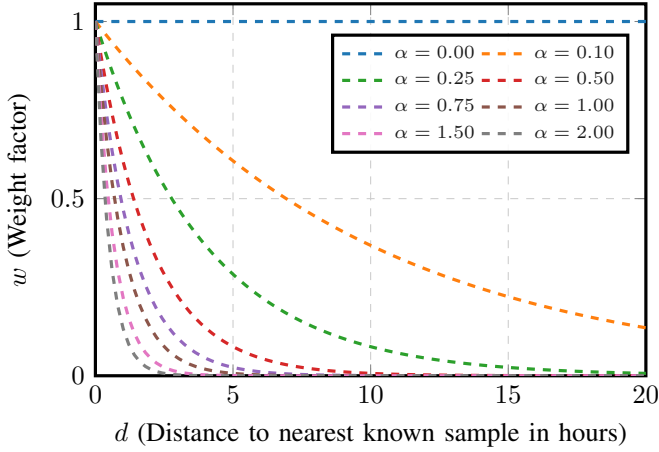

For each gap, we evaluated 12 discrete $\alpha$ candidates ranging from $0.0$ to $2.0$. Notably, this candidate pool included $\alpha = 0.1081$, the specific optimal value identified in the original OWA study \cite{peppanen_imputation}. The search was capped at $2.0$ because, as Fig.~\ref{fig:owa_weights} illustrates, the exponential weight factor $w_i$ decays to near-zero very rapidly. Any parameter beyond $0.50$ eliminates the linear interpolation component, forcing the algorithm to rely entirely on the historical average.

This randomized simulation generated millions of distinct evaluations, ensuring statistically robust parameters. The optimal $\alpha$ that minimized the Mean Absolute Error (MAE) for each gap size is presented in Table \ref{tab:owa_alpha}.

\begin{table}[!b]
\vspace{-3mm}
\caption{Optimal $\alpha$ by Gap Length}
\label{tab:owa_alpha}
\centering
\begin{tabular}{|c|c|c|c|}
\hline
\textbf{Gap Size ($G$)} & \textbf{Optimal $\alpha$} & \textbf{Evaluated Hours} & \textbf{MAE (kWh)} \\
\hline
1 hour & 0.0000 & 30,000 & 0.186 \\
\hline
2 hours & 0.0000 & 60,000 & 0.228 \\
\hline
3 hours & 0.1081 & 90,000 & 0.262 \\
\hline
4 hours & 0.2500 & 119,998 & 0.286 \\
\hline
6 hours & 0.2500 & 180,000 & 0.320 \\
\hline
8 hours & 0.5000 & 240,000 & 0.334 \\
\hline
12 hours & 0.5000 & 359,999 & 0.354 \\
\hline
24 hours & 0.5000 & 719,997 & 0.356 \\
\hline
48 hours & 0.5000 & 1,439,989 & 0.363 \\
\hline
72 hours & 0.5000 & 2,159,991 & 0.365 \\
\hline
168 hours & 0.5000 & 5,039,987 & 0.369 \\
\hline
\end{tabular}
\end{table}

The optimization results demonstrate that for micro gaps ($G \le 2$ hours), the optimal $\alpha$ is strictly $0.00$, indicating that historical data only introduce noise and pure linear interpolation is superior. As the gap widens, local edges lose predictive power, and $\alpha$ steadily increases. For all extended outages ($G \ge 8$ hours), the parameter stabilizes at $\alpha = 0.50$, shifting the reliance to historical patterns. 

With the optimal $\alpha$ parameters explicitly determined on this representative subset, we proceeded with a full-scale network evaluation, described in Section~\ref{sec:experimental}.

\subsection{Low-Rank Matrix Completion (SoftImpute)}
While the OWA algorithm excels at analyzing an individual meter's historical behavior, its accuracy inevitably degrades when communication outages last for weeks, as the local temporal context is completely erased. To address these extreme structural failures, we utilize a matrix completion approach based on spatial correlation, leveraging the \textit{SoftImpute} algorithm proposed by Mazumder et al. \cite{mazumder2010}.

Rather than looking backward or forward in time for a single meter, this approach looks \textit{laterally} across the entire smart meter network. As demonstrated by Mateos and Giannakis \cite{mateos_lowrank}, spatiotemporal load profiles inherently exhibit a low-rank structure. This means that, despite the large number of commercial clients, most share a small number of basic operational schedules. This assumption is mathematically validated on the target dataset using Truncated Singular Value Decomposition (SVD), which revealed that just 10 distinct behavioral patterns can explain $97.76\%$ of the total variance across all 17,428 meters. 

Following the principles of low-rank matrix recovery, we treat the entire network's consumption history as a massive matrix $X$, where rows represent meters and columns represent hourly timestamps. To reconstruct the missing entries, SoftImpute seeks an approximation matrix $\hat{X}$ that matches the observed data while keeping the mathematical complexity (rank) as low as possible. The optimization problem is formulated as:
\begin{equation}
    \min_{\hat{X}} \frac{1}{2} \| P_\Omega(X) - P_\Omega(\hat{X}) \|_F^2 + \lambda \|\hat{X}\|_*
    \label{eq:softimpute}
\end{equation}

In this equation, the projection operator $P_\Omega(\cdot)$ ensures that the error is only calculated over the healthy, available data. The first term (the squared Frobenius norm) guarantees that our reconstructed matrix closely matches the actual known readings. The second term is the nuclear norm ($\|\cdot\|_*$), scaled by the regularization parameter $\lambda$. SoftImpute efficiently solves this by iteratively computing a soft-thresholded SVD, which shrinks the singular values toward zero \cite{mazumder2010}. As explained in \cite{mateos_lowrank}, minimizing the nuclear norm mathematically forces the solution to be "low-rank", meaning it must rely on as few shared consumption patterns as possible.

Thus, the algorithm reconstructs missing profiles by borrowing patterns from healthy meters with similar behavior.

Accordingly, the entire dataset was converted into a global matrix of $17,428 \times 18,264$ on which this algorithm was validated according to the method explained in Section~\ref{sec:experimental}.

\subsection{Shape Modeling with Autoencoders}
The final imputation strategy evaluated in this study approaches the reconstruction problem via shape modeling. Based on the framework proposed by Duarte et al. \cite{duarte_autoencoder}, this approach argues that daily consumption profiles can be highly accurately reconstructed by separating the \textit{shape} of the daily consumption from its \textit{magnitude} (total daily consumption).

The core intuition is that commercial meters follow calendar-dependent shapes, which may differ substantially between workdays, weekends, and holidays, regardless of total daily consumption.

To implement this, we first enriched our dataset by tagging every 24-hour sequence with precise calendar features: the Day of the Year ($DOY$), mapped continuously from $0.0$ to $1.0$, and the Type of Day ($TOD$), categorized simply into workdays, Saturdays, and Sundays/national holidays. Following the data preparation protocols in \cite{duarte_autoencoder}, we isolated only the "healthy" days (days with 24 valid hourly readings), resulting in a clean training dataset of over 13 million individual 24-hour profiles. Each daily profile was then normalized by its total daily energy, transforming raw consumption values into a pure 24-hour \textit{shape}.

\begin{figure}[!b]
    \centering
    \resizebox{\columnwidth}{!}{
    \begin{tikzpicture}[
        layer/.style={draw, thick, rounded corners=2pt, align=center, text width=0.8cm},
        arrow/.style={->, >=stealth, thick, draw=gray!80}
    ]

    \node[layer, fill=cyan!10, minimum height=4.8cm] (l1) {24};
    \node[layer, fill=cyan!20, minimum height=3.2cm, right=0.6cm of l1] (l2) {16};
    \node[layer, fill=cyan!30, minimum height=1.6cm, right=0.6cm of l2] (l3) {8};

    \node[layer, fill=orange!30, minimum height=0.8cm, right=1cm of l3] (latent) {\textbf{2}};

    \node[layer, fill=green!30, minimum height=1.6cm, right=1cm of latent] (l4) {8};
    \node[layer, fill=green!20, minimum height=3.2cm, right=0.6cm of l4] (l5) {16};
    \node[layer, fill=green!10, minimum height=4.8cm, right=0.6cm of l5] (l6) {24};

    \draw[arrow] (l1) -- (l2);
    \draw[arrow] (l2) -- (l3);
    \draw[arrow] (l3) -- (latent);
    \draw[arrow] (latent) -- (l4);
    \draw[arrow] (l4) -- (l5);
    \draw[arrow] (l5) -- (l6);

    \node[below=0.2cm of l1, font=\footnotesize, align=center] {Input\\Profile};
    \node[below=0.2cm of latent, font=\footnotesize, align=center] {Latent\\Space};
    \node[below=0.2cm of l6, font=\footnotesize, align=center] {Output\\Profile};

    \draw [decorate,decoration={brace,amplitude=8pt,raise=5pt}, thick]
        ([xshift=-2pt,yshift=5pt]l1.north west) -- ([xshift=2pt,yshift=5pt]l3.north east)
        node [midway, yshift=20pt] {\textbf{Encoder}};

    \draw [decorate,decoration={brace,amplitude=8pt,raise=5pt}, thick]
        ([xshift=-2pt,yshift=5pt]l4.north west) -- ([xshift=2pt,yshift=5pt]l6.north east)
        node [midway, yshift=20pt] {\textbf{Decoder}};

    \end{tikzpicture}
    } 
    \caption{Architecture of the Autoencoder-based Neural Network for daily shape compression and reconstruction.}
    \label{fig:autoencoder_arch}
\end{figure}
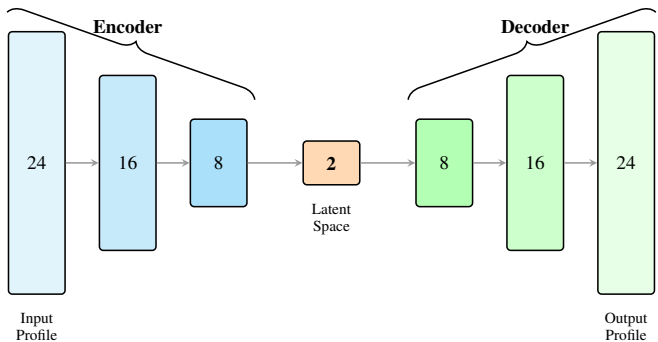

These normalized daily shapes were fed into an Autoencoder-based Neural Network. The Autoencoder's task is to compress the complex 24-hour shape through successive hidden layers (reducing dimensionality from 24 to 16, and then to 8) down to a highly compressed "latent space" consisting of just two latent variables. The decoder part of the network then attempts to reconstruct the original 24-hour shape only from these two latent variables. This is illustrated in Fig.~\ref{fig:autoencoder_arch}.
\vspace{-2mm}

\section{Experimental Setup and Metrics}\label{sec:experimental}
By processing all 13 million profiles, the trained encoder successfully learned to represent any daily consumption shape in the network using only two latent numbers. In the final step of our pipeline, we established a link between the calendar and these shapes. We trained simple Linear Regression models to predict these two latent variables (as well as the total daily energy consumption) based exclusively on the calendar features ($DOY$ and $TOD$). 

In the event of an extended outage, the calendar features of the missing day are directly utilized by the regression models to estimate the expected daily energy consumption and the two latent shape variables. These latent variables are then passed through the pre-trained Decoder network, which expands them back into a full 24-hour normalized shape. Finally, multiplying this shape by the predicted daily energy yields the fully reconstructed 24-hour consumption profile for the missing day. For gaps shorter than 24 hours, the decoder reconstructs the complete 24-hour daily profile, but only the artificially masked hours are included in the evaluation metrics.

The evaluation of this model, involving millions of simulated data gaps distributed across the entire 17,428 meter network, is detailed in the subsequent experimental setup section.

\subsection{Missing Data Simulation}
To objectively evaluate the imputation algorithms, a Monte Carlo masking strategy was utilized. Entirely healthy periods within the dataset were isolated, and continuous data gaps of 11 distinct lengths ($G \in \{1, 2, 3, 4, 6, 8, 12, 24, 48, 72, 168\}$ hours) were artificially injected. To reduce spatial and temporal bias, five independent simulation iterations were executed. In each iteration, exactly one randomly placed gap for each of the 11 sizes was introduced to every single active meter (17,428 in total). This large-scale testing ensures that the models are challenged across all possible seasonal, weekly, and daily contexts.

\subsection{Evaluation Metrics}
To quantify reconstruction accuracy, the Mean Absolute Error (MAE), the Root Mean Square Error (RMSE), and the Coefficient of Determination ($R^2$) were calculated. These metrics were computed not only globally, but also at the \textit{meter level}. Tracking the error distributions individually for all 17,428 meters enables us to visualize the true consistency, stability, and worst-case reliability of each algorithm.

Commercial meter profiles frequently exhibit near-zero nighttime consumption, causing the standard Mean Absolute Percentage Error (MAPE) to crash due to division by zero. To circumvent this mathematical instability while still capturing relative accuracy, we implemented the Weighted Absolute Percentage Error (WAPE):
\begin{equation}
WAPE = \frac{\sum_{i=1}^{N}|y_i - \hat{y}_i|}{\sum_{i=1}^{N}|y_i|} \times 100\% \label{eq:wape}
\end{equation}
where $y_i$ is the actual hidden consumption, $\hat{y}_i$ is the imputed value, and $N$ is the total number of evaluated data points.

\section{Results and Discussion}
To realistically evaluate the proposed algorithms, we analyzed the imputation performance at both a global network level and an individual meter level. In the visual representations shown below, the selected metrics reflect the average performance across the 5 independent iterations of the Monte Carlo simulation.

\subsection{Global Network Performance}
\begin{figure}[!b]
    \centering
    \includegraphics[width=\columnwidth]{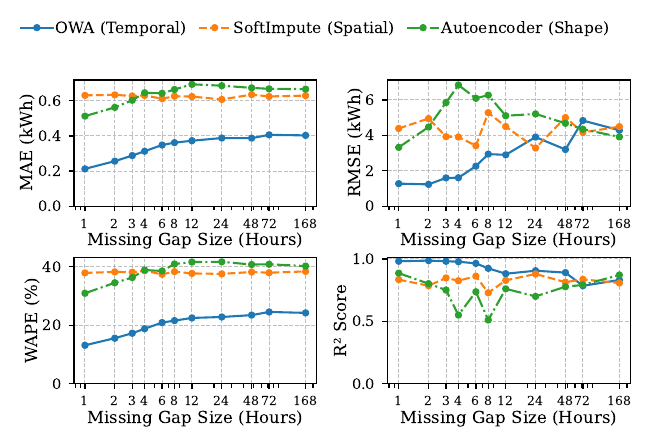}
    \vspace{-1mm}
    \caption{Global performance comparison of OWA, SoftImpute, and Autoencoder across missing gap sizes ($1$ to $168$ hours).}
    \label{fig:global_metrics}
    \vspace{-1mm}
\end{figure}

Fig.~\ref{fig:global_metrics} illustrates the performance of the three algorithms across 11 gap sizes using MAE, RMSE, WAPE, and $R^2$. The results clearly establish OWA as the most accurate global imputation method.

Consistent with the behavior reported for OWA in previous experiments \cite{peppanen_imputation}, for outages up to 24 hours the method achieves exceptionally low error and high $R^2$ by precisely interpolating from immediate neighboring data. Even for extended 168-hour gaps, the algorithm achieves the lowest absolute error by using historical data. SoftImpute serves as a robust secondary approach, keeping error rates bounded by leveraging spatial correlations from neighboring meters.

In contrast, although autoencoder-based shape modeling has been shown to produce plausible daily profiles in previous work \cite{duarte_autoencoder}, in our large-scale commercial setting it shows the greatest overall error. This outcome is expected, since it reconstructs a smoothed, generic daily profile from calendar features and therefore cannot capture sudden, irregular consumption spikes. However, its primary advantage is its completely flat error curve. Because it does not rely on adjacent hourly measurements, it is entirely immune to gap size, reconstructing a 168-hour outage with the exact same baseline accuracy as a 1-hour outage.

\subsection{Meter-Level Stability and Outliers}
\begin{figure}[!b]
    \centering
    \includegraphics[width=\columnwidth]{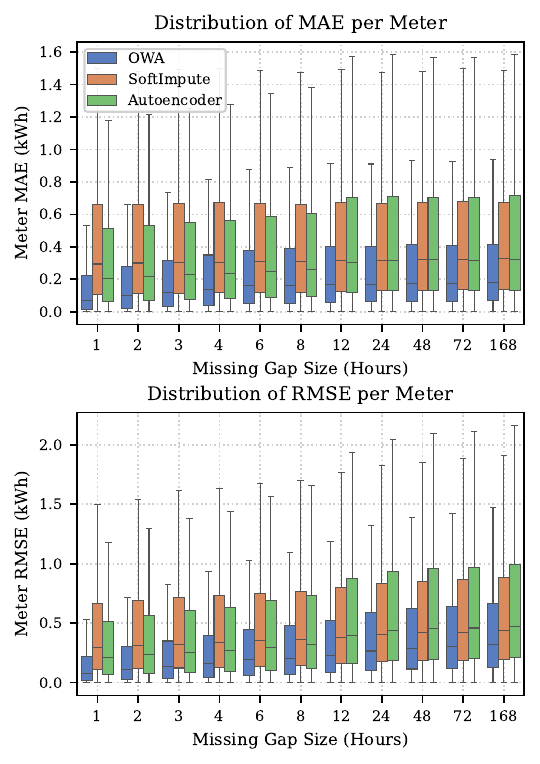}
    \vspace{-3mm}
    \caption{Meter-level distribution of MAE and RMSE.}
    \label{fig:boxplots}
\end{figure}

While global averages are useful, commercial profiles are highly volatile, requiring a closer look at individual meter stability. Fig.~\ref{fig:boxplots} presents the MAE and RMSE distributions across all 17,428 meters.

The boxplots confirm OWA's dominance at the individual meter level. Across all gap sizes, it maintains the lowest median error and the tightest interquartile range (IQR), indicating reliable reconstruction for most consumers. SoftImpute shows moderate variance, whereas the Autoencoder, though unaffected by the gap size, has a higher median error and wider spread, especially for meters with highly specific behaviors.

\section{Conclusion and Future Work}
Commercial electricity consumption profiles are highly specific and volatile, requiring specialized imputation techniques. To ensure the reliability of the findings for real-world smart grids, three distinct algorithms were evaluated on a large-scale dataset of over 17,400 active smart meters. The evaluation shows that the OWA algorithm is the most accurate method for routine data reconstruction. Since electricity usage follows strong daily and weekly patterns, OWA intelligently uses past and future anchors (e.g., matching a missing Monday with adjacent Mondays) to maintain exceptionally low error rates, even for outages lasting up to a full week. 

However, if a meter loses temporal context for an extended period, alternative approaches become necessary. Spatial models such as SoftImpute or shape-based models such as Autoencoders can serve as fallback mechanisms when temporal data are unavailable.

Future work will focus on developing an adaptive, \textit{hybrid imputation framework}. Rather than relying on a single method, this system will analyze the duration and context of each gap, dynamically routing the task to the most appropriate algorithm to ensure maximum grid resilience.

\section*{Acknowledgment}
This work has been supported by the European Union's Horizon Europe programme project INITIATE, Grant Agreement ID 101136775.
\balance
\bibliographystyle{IEEEtran}
\bibliography{references}
\end{document}